# An Automated Georeferencing Technique for Multi-Temporal Stope Point Clouds for Downstream Geotechnical Analysis

Dibyayan Patra[1,*], Simit Raval[1], Pasindu Ranasinghe[1], Bikram Banerjee[2], Ismet Canbulat[1]

[1]School of Minerals and Energy Resources Engineering, University of New South Wales, Sydney, NSW, Australia.

[2]School of Science, Engineering and Digital Technologies, University of Southern Queensland, Toowoomba, QLD, Australia.

[*]Corresponding Author Email: d.patra@unsw.edu.au

**Abstract:** The increasing use of UAV laser scanning in underground mines has enabled frequent acquisition of 3D point clouds from challenging environments such as stopes, generating large volumes of multi-temporal spatial data throughout successive stages of excavation. However, in GNSS-denied underground environments, independently acquired stope point clouds are generated within local scanner reference frames and require registration and georeferencing before they can be integrated with mine reference data and meaningfully utilised for downstream geotechnical analysis, monitoring, and mine planning. This process is commonly performed manually by aligning individual stope scans with mine reference drives, making repeated georeferencing of multi-temporal datasets time-consuming and potentially limiting the utilisation of routinely acquired data. This study proposes the 3D Tag-based Automated Registration and Georeferencing Technique (3D-TARGeT), an automated framework using low-cost, generic, non-unique rectangular georeferencing tags to establish spatial correspondence between independently acquired stope point clouds and the mine reference coordinate system. The proposed framework combines automated tag identification with geometric tag matching and rigid transformation estimation to achieve automated registration and georeferencing. The framework was evaluated as a proof of concept using four multi-temporal point-cloud scans of an underground mine stope, with the proposed georeferencing tags simulated under representative scanning conditions. 3D-TARGeT achieved consistent centimetre-level georeferencing accuracy, with median cloud-to-cloud distance and root mean square error remaining below 0.03 m across all scans, while substantially outperforming widely used automatic point-cloud registration techniques. Overall, 3D-TARGeT provides an accurate and robust approach for automating stope point-cloud georeferencing, reducing reliance on time-intensive manual alignment and facilitating the utilisation of multi-temporal datasets for downstream geological and geotechnical applications.

## 1. Introduction

The rapid advancement of three-dimensional remote sensing and laser scanning technologies has substantially increased the volume and frequency of spatial data acquired in underground mining operations. In particular, the increasing availability of compact unmanned aerial vehicle (UAV)-mounted laser scanning systems has enabled high-resolution point clouds to be collected rapidly from complex and otherwise difficult-to-access mine environments (Ellmann et al., 2021; Jones et al., 2019; Kumar Singh et al., 2023; Patra, Baylis, et al., 2025; Štroner et al., 2025). Underground stopes are large underground excavations created through successive blasting and extraction of an orebody, typically progressing through multiple stages as the broken material is removed and new rock surfaces are exposed. Stopes therefore represent an important application because their large, irregular excavation geometry, unsupported surfaces, and restricted physical accessibility make conventional direct surveying and geotechnical mapping challenging and potentially hazardous. UAV laser scanning allows these excavations to be remotely captured following successive excavation and blasting activities, resulting in increasingly large multi-temporal datasets that document the evolution of stope geometry throughout the mining cycle. Such datasets have significant potential for applications including structure mapping, stope and blast reconciliation, analysis of structural propagation, change and deformation monitoring, safety assessment, excavation and quality assurance/quality control (QA/QC), mine planning, construction performance assessment, and mine logistics (Daghigh et al., 2022; Kumar Singh et al., 2023; Monsalve et al., 2019; Walton et al., 2018; Wang et al., 2025). However, realising this potential requires the acquired point clouds to be spatially referenced within the established mine coordinate system.

This requirement presents a fundamental challenge in underground environments, where Global Navigation Satellite System (GNSS) signals are unavailable. Underground mine spatial reference frameworks are conventionally established through mine surveying, whereby surface or existing survey control is progressively transferred underground using instruments such as total stations (Bian et al., 2026; Ellmann et al., 2021; Nguyen et al., 2023). The resulting surveyed mine control provides the basis for georeferenced mine development and reference-drive geometry within the mine coordinate system. In contrast, point clouds acquired using UAV-mounted and simultaneous localisation and mapping (SLAM)-based laser scanning systems are generally generated within a local scanner reference frame (Puente et al., 2013; Raval et al., 2019). Consequently, a newly acquired stope point cloud cannot be directly spatially integrated with the corresponding mine geometry solely on the basis of its local coordinates and must first be registered and georeferenced to the established mine reference framework. This spatial correspondence is essential not only for locating an individual scan correctly within the mine, but also for integrating multiple scans acquired at different stages of excavation and enabling meaningful spatial, geotechnical, and temporal analysis.

The current industry-standard practice for georeferencing stope point clouds is predominantly a manual process undertaken by mine surveyors and geotechnical engineers (Walker & Awange, 2020). Newly acquired stope scans are manually aligned with georeferenced mine reference drives by identifying common access-drive and surrounding tunnel geometry and iteratively translating and rotating the point cloud until satisfactory spatial alignment is achieved. While this provides a practical means of bringing individual scans into the mine coordinate system, the process is significantly time-consuming and dependent on operator interpretation. The problem becomes increasingly significant with multi-temporal monitoring, where the procedure must be repeated independently for each newly acquired scan. The growing efficiency of UAV laser scanning therefore creates a processing imbalance. Large quantities of spatial data can be acquired relatively quickly and routinely, while the initial georeferencing step remains comparatively labour-intensive. As the number and frequency of scans increase, this can create a workflow backlog in which potentially valuable datasets remain underutilised because substantial manual processing is required before downstream analysis can even begin. In this sense, increasing data acquisition capability alone does not necessarily translate into increasing geotechnical information unless the corresponding spatial-processing workflow can also be automated.

Automatic point-cloud registration provides a potential means of addressing this limitation. Widely established approaches include the Iterative Closest Point (ICP) algorithm and Normal Distributions Transform (NDT), which estimate transformations from the geometric relationship between point-cloud datasets (Dong et al., 2020; Magnusson et al., 2009; Pang et al., 2018; Pomerleau et al., 2015; Sun et al., 2024; Yang et al., 2021). However, their performance depends strongly on the availability of sufficient common and distinguishable geometry. Underground mine environments present several difficulties in this regard, including repetitive tunnel geometry, geometrical symmetry, and limited distinctive features, which can introduce ambiguous or false geometric correspondences. The problem is further amplified for stope georeferencing because progressive excavation substantially changes the scanned geometry between acquisition stages, while the reference-drive dataset represents only a comparatively small portion of the geometry contained within the complete stope scan. As a result, the large, excavated stope surfaces may have no corresponding geometry within the reference dataset, leaving only limited tunnel access-drive regions available for geometry-based registration. These characteristics make conventional geometry-based registration particularly challenging for stope georeferencing, where the transformation must be recovered from comparatively limited and continuously changing common geometry.

Ground control targets (GCTs), on the other hand, are increasingly being incorporated into mine surveying and spatial monitoring operations to provide reliable reference features for point-cloud registration, georeferencing, calibration, localisation, and drift correction (Becerik-Gerber et al., 2011; Kumar Singh et al., 2023; Lavigne & Marshall, 2012; Makuch & Gawronek, 2025; Muralikrishnan, 2021; Urbancic et al., 2019). These targets can broadly be classified as active or passive. Active GCTs, such as RFID or wireless sensor-based systems, can provide identifiable spatial references but typically require additional power, sensing or communication hardware, and associated installation and system integration. Passive GCTs, in contrast, require no onboard power and can be identified from their optical, intensity, or geometric characteristics, making them a relatively simple and practical option for underground surveying applications. The absolute coordinates of installed GCTs can also be established within the mine coordinate system using conventional total-station surveying, allowing them to act as known spatial references for subsequently acquired point clouds. Passive GCTs can therefore be applied for various purposes including spatial referencing, scanner calibration, data coregistration, and reduction of accumulated mapping drift.

However, multi-temporal stope georeferencing presents a distinctly different challenge from conventional point-cloud co-registration, where successive datasets generally represent substantially the same pre-existing environment. A stope is an actively evolving excavation. Successive blasting and material extraction progressively create new excavation volumes and expose rock surfaces that did not exist in the preceding scan or the original mine reference geometry. Consequently, a large proportion of each newly acquired stope point cloud may have no physical counterpart in either earlier scans or the established reference drives, while the common geometry may be restricted primarily to relatively small access-drive regions. This combination of evolving geometry, restricted physical access, limited common reference features, and repeated multi-temporal acquisition makes automated stope georeferencing a particularly challenging spatial referencing problem. Drawing inspiration from the concept of passive GCT-based spatial referencing, artificial geometric references offer a potential means of overcoming the dependence on common natural geometry in such dynamically evolving environments.

Building on this concept, this study proposes the 3D Tag-based Automated Registration and Georeferencing Technique (3D-TARGeT), specifically developed for automated georeferencing of multi-temporal stope point clouds. The framework introduces simple, low-cost, generic and non-unique rectangular geometric tags that can be automatically identified directly from the 3D point cloud. Rather than requiring individually encoded targets, 3D-TARGeT resolves tag correspondence using their relative spatial arrangement and uses the resulting matched locations to establish spatial correspondence between the locally referenced stope scan and the georeferenced mine reference drives. Rigid transformation estimation and subsequent refinement are then used to transform the complete stope point cloud into the established mine coordinate system. The proposed framework is evaluated as a proof of concept using real multi-temporal point-cloud scans acquired from an underground mine stope. In this proof-of-concept study, the proposed tags were simulated and incorporated into the acquired datasets, with representative tag geometry, point density, and scanner characteristics considered to reproduce practical scanning conditions. By introducing persistent geometric references specifically tailored for the application, 3D-TARGeT provides an automated georeferencing approach that reduces the time and effort associated with repeated manual georeferencing, thereby improving the availability and usability of multi-temporal stope point-cloud datasets for downstream geological and geotechnical analysis.

The remainder of this paper is organised as follows. Section 2 presents the materials and methodologies utilised in the study, including the study area and georeferencing setup, followed by the complete 3D-TARGeT methodology comprising data filtering, tag identification, registration and georeferencing, validation, and the downstream structure-mapping framework. Section 3 presents and discusses the tag-identification and georeferencing results, evaluates the accuracy and performance of 3D-TARGeT against conventional registration techniques, and demonstrates its application to multi-temporal structure mapping. Section 4 outlines future research directions and potential extensions of the proposed framework, while Section 5 summarises the key findings and conclusions of the study.

## 2. Materials and Methods

### 2.1. Study Area and Georeferencing Setup

For this study, a multi-temporal point cloud dataset of an underground mine stope was used. The stope is located within an active operational underground metal mine. In the context of this study, multi-temporal refers not only to the time progression of the stope but also to its vertical and volumetric progression throughout the mining sequence. Stopes are large excavated chambers created during underground ore extraction. By their nature, stopes are developed through multiple planned blast packets rather than a single excavation. These blast packets are carried out over time, with each successive blast progressively extending the excavation to create a larger stope volume. After each blast packet, although the excavation progressively extends upward, increasing the overall stope volume, some access points at the lower levels become blocked by blasted debris and localised structural failures. These access points are typically reopened only after the blasted material has been extracted. The stope used in this study is a multi-lift sublevel open stope. Following each blast packet, the stope is accessible from three or more access points and is generally accessible from the upper crown level and, once mucking operations commence, from the lower draw points. Consequently, the available scanning locations and the visible portions of the excavation evolve throughout the mining sequence, resulting in a challenging multi-temporal point cloud dataset with continuously changing geometry, visibility, and scan overlap. The multi-temporal dataset used in this study is illustrated in Figure 1, showing the progressive evolution of the stope over successive blast packets.

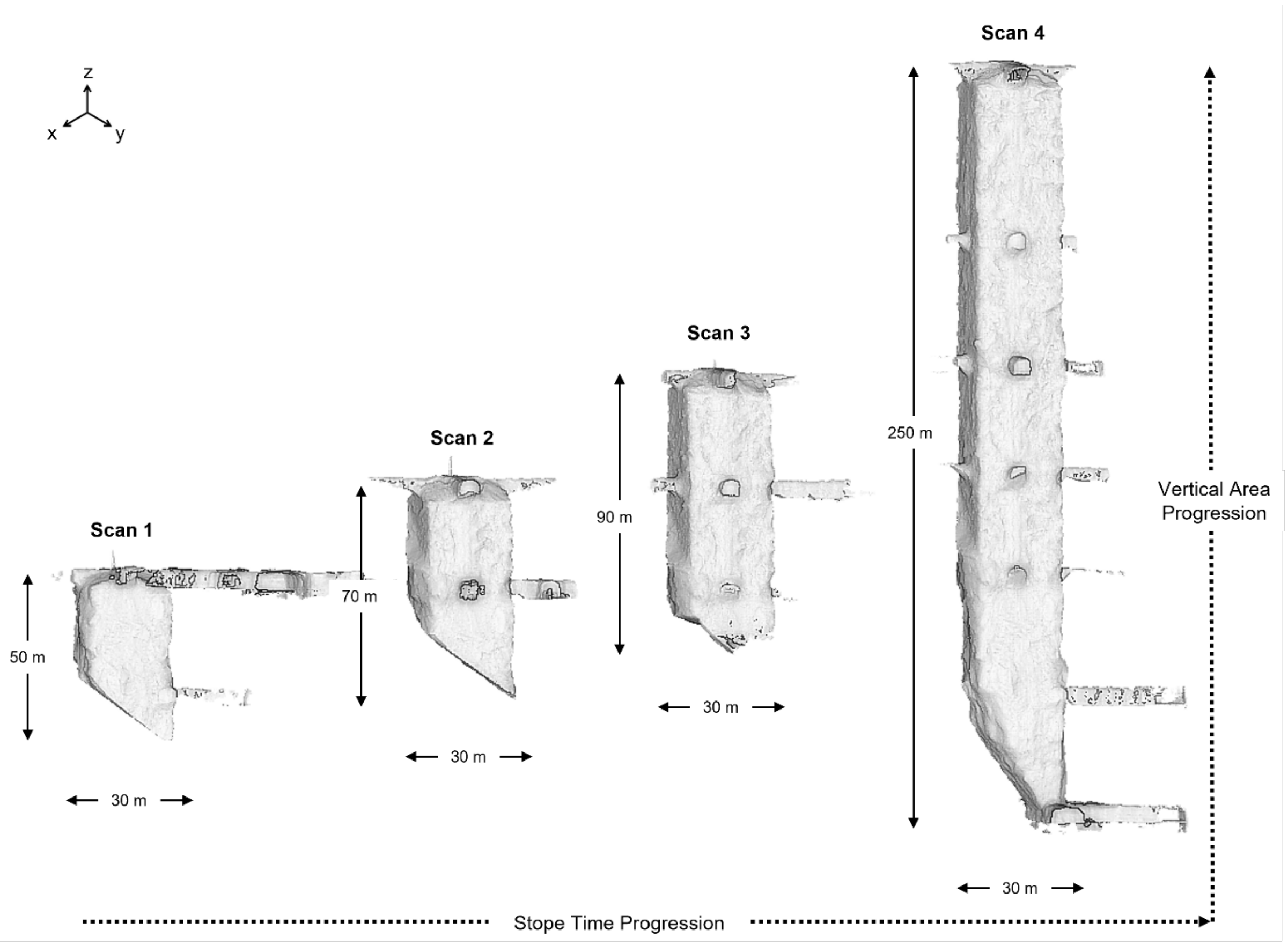

***Figure 1.*** *Multi-temporal progression of the study stope showing four representative point cloud scans acquired at different stages of excavation. The marked approximate dimensions illustrate the progressive increase in stope vertical extent, while maintaining an approximately constant plan width.*

These scans are collected using an Emesent Hovermap scanner comprising a Velodyne 16-channel LiDAR mounted on a rotating head. The device is mounted on a DJI M210 Enterprise UAV and flown within the stope cavity. The system uses Wildcat SLAM, originally developed by CSIRO, to co-register and process the raw LiDAR scans together with the embedded IMU telemetry data into a high-resolution 3D point cloud of the underground environment. In mine stope environments, direct GNSS-based georeferencing of independently acquired point clouds is not possible due to the lack of GNSS availability. Therefore, in this study, low-cost, non-unique rectangular identifiers or georeferencing tags are proposed for the automated georeferencing of multi-temporal stope point clouds.

The dimensions of the proposed tag are dictated by scanner characteristics, particularly the laser beam divergence and the resulting point density of the generated point cloud. The scanner produced point clouds with an approximate point density of 800 points/m². The horizontal and vertical beam divergence of the LiDAR scanner used in the Hovermap system is illustrated in Figure 2. Beam divergence determines the physical laser footprint, or maximum ground field of view, of an individual laser pulse on the scanning surface, with the footprint increasing as the distance between the scanner and surface increases (Centeno et al., 2010; Kaasalainen et al., 2011; Kellner et al., 2019; Ranasinghe et al., 2026; Sabzali & Pilgrim, 2025). Importantly, beam divergence does not directly represent point spacing or point density; rather, it defines the area illuminated by an individual laser pulse. Consequently, when a feature approaches or becomes smaller than the laser footprint, the return may represent a mixture of the feature and its surrounding surface, reducing the reliability with which its geometry and boundaries can be resolved. Figure 2 also presents the corresponding maximum horizontal and vertical ground fields of view at different scanner-to-surface distances, illustrating the progressive increase in laser footprint with distance. Although the mobile scanning process generally observes a surface multiple times from different positions and SLAM integrates these observations to increase the effective spatial sampling density, the physical footprint of each individual laser pulse remains unchanged. In cases where a region is observed only briefly due to factors such as the UAV flight path, sudden motion, limited visibility, or sensor orientation, the beam footprint therefore remains an important consideration when determining the minimum dimensions of the georeferencing tag. Using the basic trigonometric relationship presented in Equation 1, the effective laser footprint at a sensor-to-target distance of 40 m was calculated based on the horizontal and vertical beam divergence. A distance of 40 m was considered as a conservative maximum scanner-to-tag distance based on the dimensions of the investigated stope. As illustrated in Figure 2, increasing beam divergence and scanning distance increases the physical area illuminated by an individual laser pulse, thereby reducing the ability to reliably resolve surface features approaching the dimensions of the laser footprint. Therefore, the proposed tag must be sufficiently larger than the corresponding laser footprint and provide adequate surface area to capture multiple LiDAR returns for reliable identification within the point cloud.

$$FoV = 2 \times d \times tan\left(\frac{\theta}{2}\right) \tag{1}$$

*Where FoV is Maximum Ground Field of View, d is the distance of the scanning surface, and θ is the horizontal/vertical beam divergence of the laser scanner.*

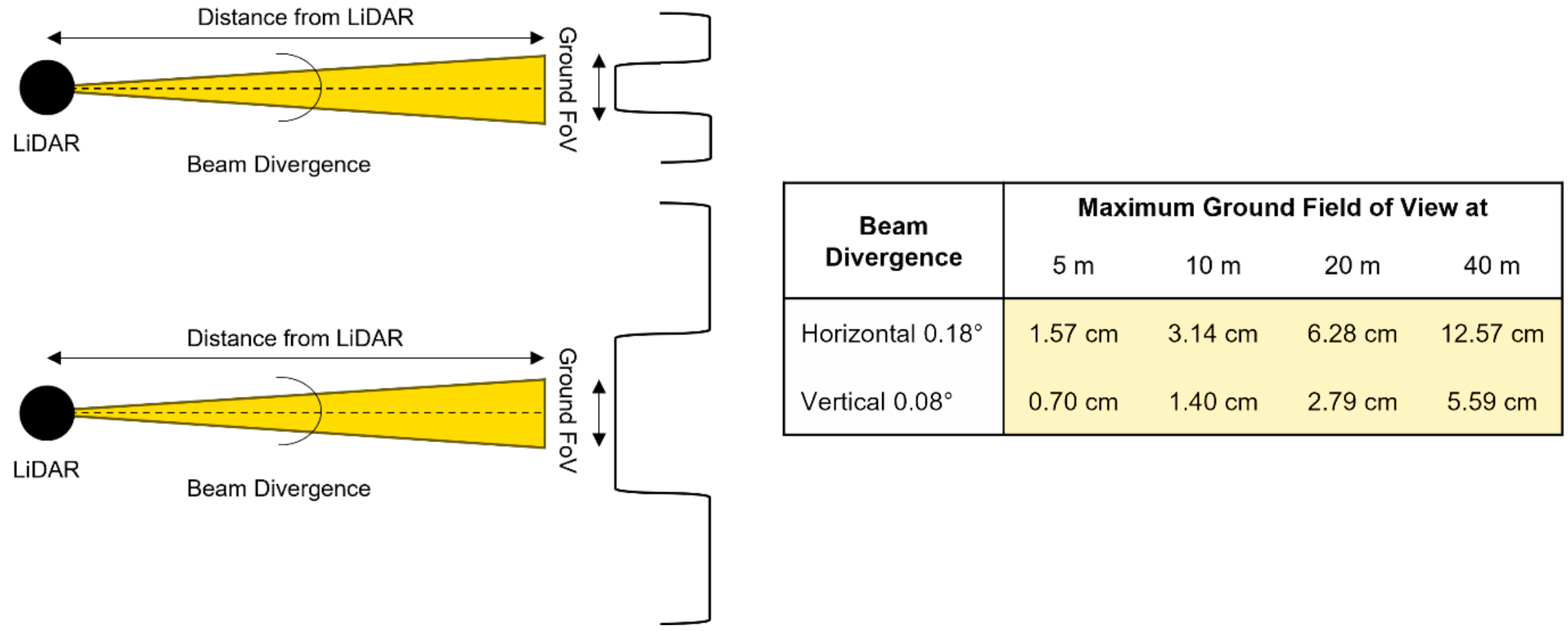

| Beam Divergence | Maximum Ground Field of View at | | | |
|---|---|---|---|---|
| | 5 m | 10 m | 20 m | 40 m |
| Horizontal 0.18° | 1.57 cm | 3.14 cm | 6.28 cm | 12.57 cm |
| Vertical 0.08° | 0.70 cm | 1.40 cm | 2.79 cm | 5.59 cm |

***Figure 2.*** *Effect of LiDAR static beam divergence on the resulting laser footprint at different scanning distances.*

Based on the beam-divergence calculations and nominal point density of the acquired point clouds, the georeferencing tag dimensions were empirically selected as 60 cm × 40 cm. These dimensions provide a sufficiently large target relative to the maximum laser footprint considered in the preceding analysis while remaining compact enough to avoid obstructing access within the mine tunnels. At the nominal point density of approximately 800 points/m², a tag surface area of 0.24 m² would contain approximately 192 points, providing sufficient spatial sampling for subsequent automated identification. For practical implementation, the tag can be manufactured from a rigid material with a matte surface finish, such as anodised aluminium or matte rigid PVC/acrylic, as matte and rough surfaces generally promote diffuse laser backscatter while reducing undesirable specular reflections (Haner & Menzies, 1989; Koch et al., 2017; Tan & Cheng, 2017; Yuan et al., 2020). For this proof-of-concept study, the tags were simulated since the investigated site is an active operational mine that has progressed beyond its development phase, and the point cloud dataset was acquired prior to the development of the proposed approach, making retrospective installation of physical tags at the corresponding stope access locations impractical. A 3D model of the generic, non-unique rectangular tag was created in SolidWorks with the specified dimensions and a nominal thickness of 2 cm. The tag was designed to hang from the tunnel roof using 60 cm suspension elements, providing sufficient separation from the roof and surrounding mine infrastructure, such as rock bolts and ventilation services, to maintain clear scanner visibility. Tag placement requires only limited consideration. The tag should preferably face towards the stope access to maximise its visibility during scanning from inside the stope and be positioned within approximately 10 m of the access point, thereby remaining comfortably within the conservative 40 m scanner-to-tag distance considered in the beam-divergence analysis. Within this 10 m region, the precise tag position does not need to be predefined, and moderate deviations from a directly facing orientation can be accommodated, enabling relatively flexible and low-effort field installation.

Figure 3a shows the georeferenced reference mine tunnel drives together with the as-planned final stope design. The reference drives were established within the mine coordinate system through total-station surveys tied to surface control, with surveyed targets subsequently identified within the mine scans to establish spatial correspondence. These georeferenced drives provide the spatial reference currently used to manually georeference the stope point clouds acquired during the mining sequence. Manual georeferencing is performed by iteratively translating and rotating each captured point cloud until the visible tunnel accesses and surrounding tunnel geometry align with the corresponding reference drives. Although this provides a practical means of establishing the scan position within the mine coordinate system, the procedure is time-consuming and dependent on manual interpretation, motivating the development of an automated alternative. For the proof-of-concept simulation, the modelled tags were synthetically positioned within the reference drives according to the placement considerations described above. An example is shown in Figure 3b, where a simulated tag, highlighted in green, is suspended from the tunnel roof within a tunnel providing access to the stope. The study assumes that the positions of these installed tags are surveyed and known within the same coordinate system as the reference drives, thereby allowing them to act as spatial reference markers. The corresponding locations were then identified within the independently acquired stope point clouds, and simulated tag point clouds were inserted at these locations, as illustrated in Figure 3c. To reproduce the discrete nature of a LiDAR-observed surface, the three-dimensional tag models were converted to point clouds using the CloudCompare Random Point Sampling tool at the nominal stope point density of 800 points/m². Random surface sampling was intentionally adopted rather than explicitly defining the tag edges and vertices, producing irregular boundary sampling more representative of real LiDAR observations and preventing the subsequent identification algorithm from relying on perfectly defined synthetic edges. The resulting dataset therefore consists of stope point clouds and georeferenced reference drives containing corresponding simulated generic georeferencing rectangular tags, providing a controlled proof-of-concept dataset for evaluating the proposed 3D Tag-based Automated Registration and Georeferencing Technique (3D-TARGeT).

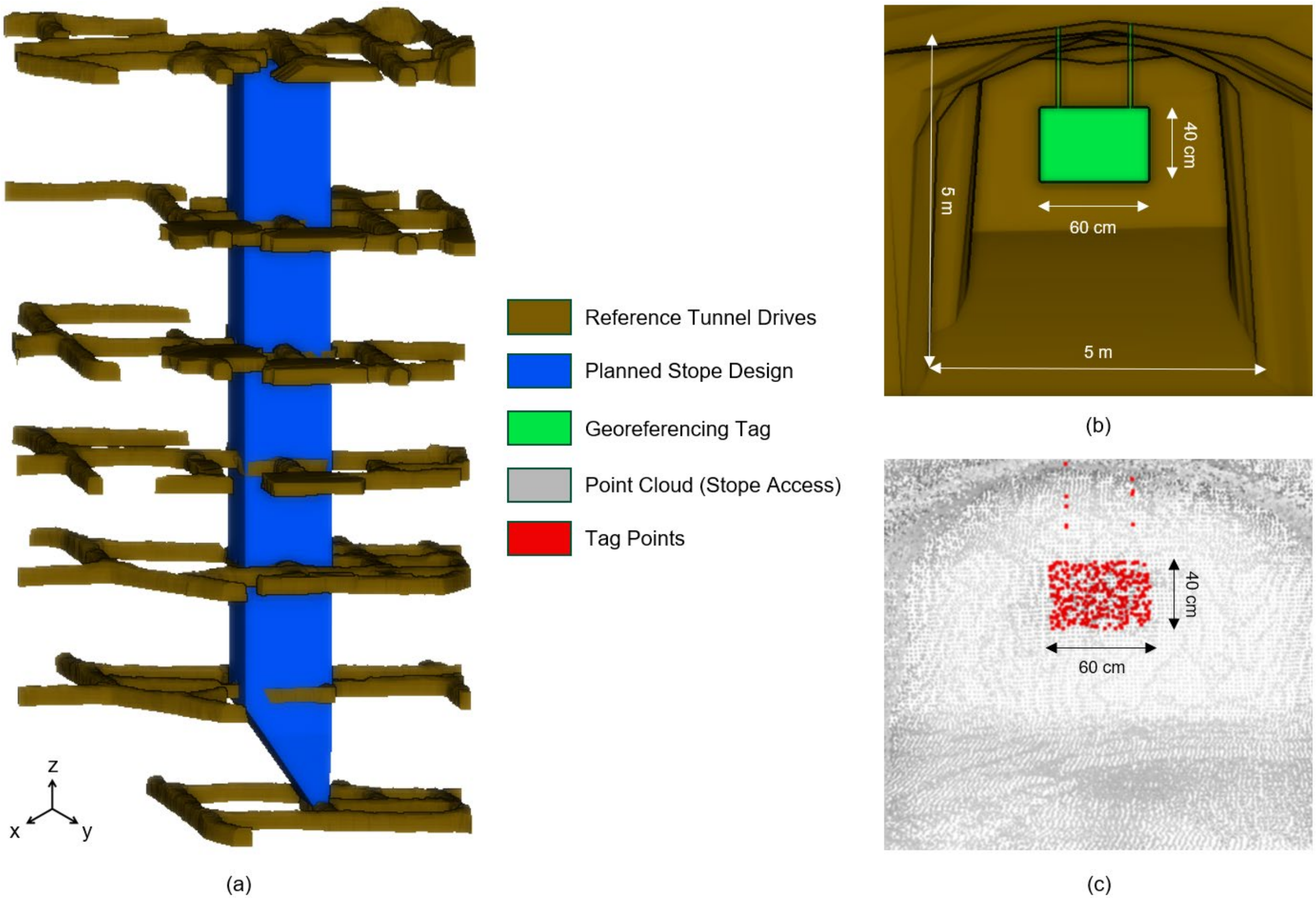

***Figure 3.*** *(a) Mine reference tunnel drives used as the reference framework for manual georeferencing of stope point clouds, alongside the as-planned stope design. (b) Enlarged view of a stope access point showing the placement and dimensions of the georeferencing tag within the reference drive. (c) Corresponding point cloud view showing the simulated georeferencing tag points used for automated georeferencing of multi-temporal stope point clouds.*

## 2.2. Overview of 3D-TARGeT

A detailed overview of the complete 3D-TARGeT processing workflow is presented in Figure 4. The workflow begins by passing each multi-temporal stope point cloud through a filtering stage to remove noise, outliers, and spurious points from the acquired data. Following filtering, a tailored tag identification procedure is applied to robustly segment the generic georeferencing tags present within each scan and determine their corresponding rectangular centres and corners. The identified tags are subsequently used together with the reference tags during the registration and georeferencing stage to transform each stope point cloud from its local scanner reference frame into the global mine coordinate system. Applying this procedure across all multi-temporal scans produces spatially registered and georeferenced datasets that can subsequently support downstream geotechnical analyses by engineering personnel for geological and geotechnical interpretation, ground understanding, decision-making, and mine planning. Each component of the proposed 3D-TARGeT workflow is described in detail in Sections 2.3 - 2.5.

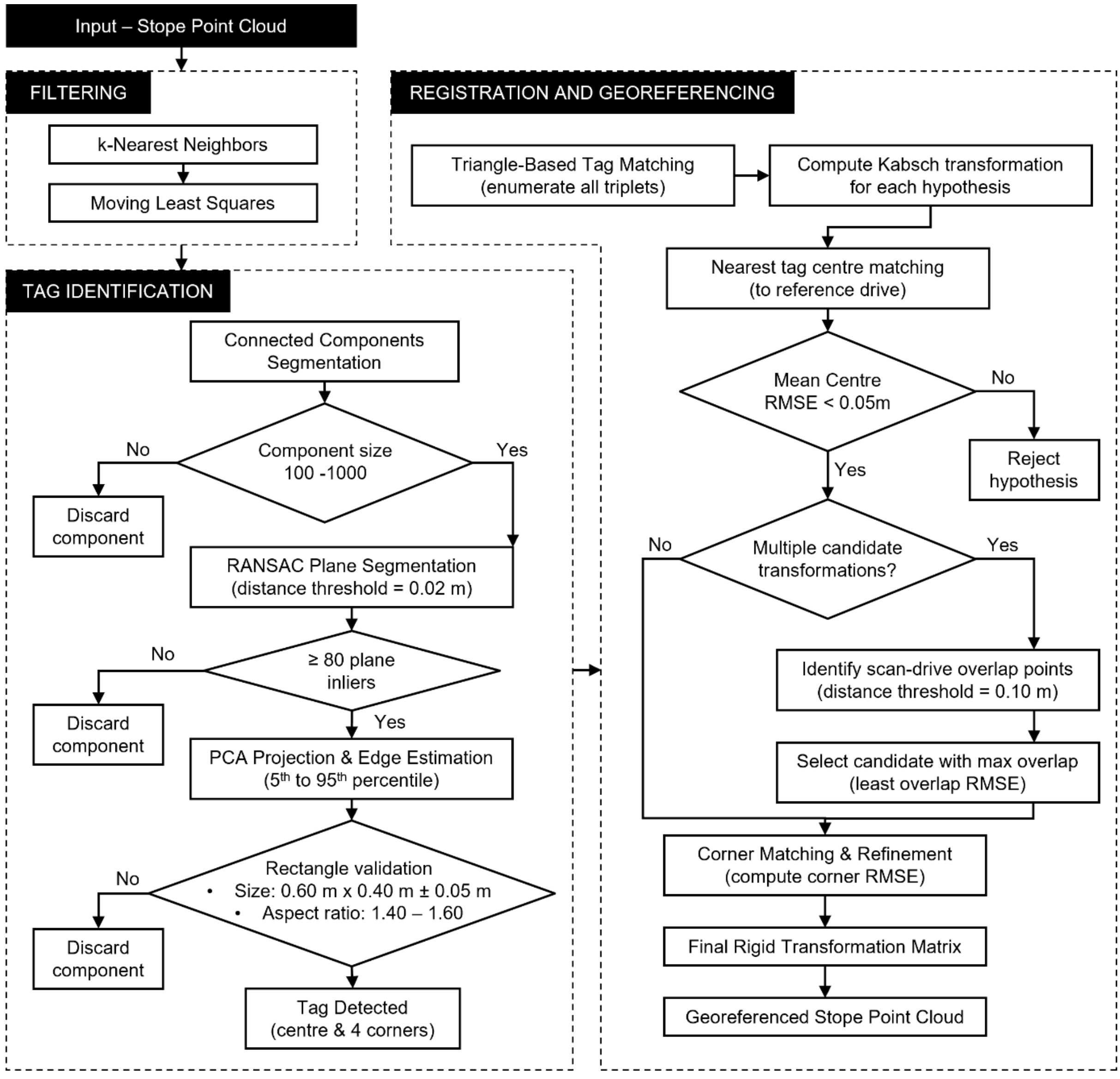

***Figure 4.*** *A detailed overview of the 3D-TARGeT algorithm used in this study.*

### 2.3. Data Filtering

A data pre-processing or filtering stage is applied to reduce erroneous, isolated, and noisy points within the acquired tagged multi-temporal stope point clouds prior to tag identification and georeferencing. Such points can arise from several factors associated with LiDAR data acquisition in complex underground environments, including beam divergence, surface reflectivity, multipath effects, and sensor perturbations. The presence of these points can introduce local surface irregularities. To address these effects, standard and well-established point cloud filtering techniques widely utilised for LiDAR data preprocessing were adopted in this study, consisting of k-nearest neighbour (k-NN) statistical outlier removal followed by Moving Least Squares (MLS) smoothing (Alexa et al., 2003; Han et al., 2017; Patra, Ranasinghe, et al., 2025a; Sankaranarayanan et al., 2007). The k-NN filter evaluates each point based on the spatial distribution of its neighbouring points and identifies points with anomalously large neighbour distances as outliers. A neighbourhood size k of 6 and a standard-deviation multiplier of 1.0 were used in this study. The relatively small neighbourhood was selected to maintain a localised assessment of point distribution, thereby removing isolated points while limiting unnecessary alteration of genuine geometric features and the comparatively small georeferencing tags. Following outlier removal, MLS filtering was applied using a first-order polynomial approximation with a 0.05 m search radius. The first-order polynomial provides a locally linear surface approximation suitable for reducing small-scale measurement irregularities without imposing higher-order surface behaviour, while the 0.05 m neighbourhood provides sufficient local support for suppressing small-scale measurement irregularities without substantially smoothing larger geometric features. Together, the two filtering stages provide complementary functions where the k-NN filter removes isolated and statistically inconsistent points, and MLS regularises the remaining local point distribution and reduces surface noise. The combination of k-NN outlier removal and local MLS smoothing therefore provides a conservative preprocessing step for the 3D-TARGeT algorithm that reduces point-cloud noise while preserving the geometric characteristics.

### 2.4. Tag Identification

Following filtering, the tag identification stage of 3D-TARGeT is applied to automatically detect the generic rectangular georeferencing tags within each stope point cloud and extract their centres and four corners for subsequent registration and georeferencing. As shown in Figure 4, the procedure consists of connected-component (CC) segmentation, RANSAC plane segmentation, and PCA-based projection and edge estimation, followed by geometric validation of the candidate components.

First, connected components (CC) segmentation is applied to separate spatially connected groups of points into individual components. CC segmentation groups neighbouring points that are connected within the discretised 3D space while assigning spatially disconnected groups to separate components. CC segmentation was selected over conventional distance- or density-based clustering approaches because the suspended tags are expected to occur as near spatially isolated objects, allowing candidate components to be directly separated based on spatial connectivity without requiring assumptions regarding cluster density or the number of clusters. This provides an efficient first-stage segmentation because the tags are designed to hang below the tunnel roof and therefore occur predominantly as spatially isolated objects within the scan. An octree level of 11, corresponding to a grid step size of approximately 0.045 m, was empirically selected to provide a suitable balance between spatial resolution and computational efficiency, retaining sufficient geometric detail for tag separation without unnecessarily increasing processing time.

Following segmentation, only components containing 100 to 1000 points are retained for subsequent processing. The relatively broad range was selected to accommodate variations expected under practical scanning conditions. In particular, genuine tags may contain fewer points when viewed from an unfavourable scanning angle or sensor trajectory, while some components may contain additional points arising from residual noise or nearby miscellaneous mine features. An example of the resulting CC segmentation is shown in Figure 5a, illustrating a candidate tag component containing inherent noise, suspension points and poorly defined edges, alongside a non-tag component representing a random stray feature within the mine environment. Components containing fewer than 100 points are therefore considered insufficiently populated and are discarded as small noise or stray-object components, whereas components exceeding 1000 points are excluded as they are unlikely to represent the relatively small georeferencing tags and would introduce unnecessary computational processing in the subsequent stages.

Each retained component is subsequently subjected to Random Sample Consensus (RANSAC) plane segmentation to determine whether it contains the dominant planar surface expected from the rectangular georeferencing tag. RANSAC is a robust model-fitting algorithm that iteratively selects subsets of points to estimate candidate plane models and evaluates each model according to the number of points that lie within a specified perpendicular distance from the estimated plane. The model receiving the greatest support is retained, with the points satisfying the distance criterion classified as plane inliers. RANSAC was selected over conventional least-squares plane fitting because its consensus-based formulation is less sensitive to outliers, enabling the dominant planar tag surface to be robustly extracted even when the connected component contains suspension elements, residual noise, or other non-planar points. In this study, a point-to-plane distance threshold of 0.02 m is used to accommodate local measurement noise, small surface irregularities, and tag thickness while maintaining a sufficiently restrictive tolerance for isolating the planar tag surface. Figure 5b illustrates this process for a representative candidate component, showing the fitted RANSAC plane, the retained planar inliers and the rejected non-planar outliers. This step is particularly important under practical scanning conditions, as a connected component containing a tag may also incorporate points associated with the suspension elements, residual noise, or other small nearby features; these points are not necessarily coplanar with the principal tag face and are therefore excluded from the subsequent geometric analysis.

To ensure that the fitted plane is supported by a sufficient number of observations, a minimum of 80 RANSAC plane inliers is required for a component to progress to the subsequent PCA-based geometric analysis. This threshold corresponds to four-fifths of the minimum retained component size, ensuring that even the smallest accepted components contain a substantial number of points supporting the planar tag surface. At the same time, the criterion provides sufficient allowance for points associated with the suspension elements, residual noise, or other non-planar features that may have been grouped with the tag during connected-component segmentation. Components producing fewer than 80 planar inliers are therefore discarded, as the fitted planar region is considered insufficiently populated to reliably represent a tag and may instead correspond to incidental near-planar surfaces within other components. Importantly, the criterion is based on the absolute number of planar inliers rather than a fixed planar-inlier proportion for every component, allowing genuine tags containing additional non-planar points to remain detectable provided that a sufficiently well-supported planar tag face can be isolated. The resulting RANSAC inliers therefore represent a potential candidate tag surface.

The extracted planar points are then processed using Principal Component Analysis (PCA) to establish the two dominant directions of variation within the fitted plane. PCA transforms the arbitrarily oriented 3D planar points into a local coordinate system aligned with the principal axes of the candidate rectangular tag, allowing its length, width, centre, and boundary locations to be estimated independently of its orientation within the mine. To reduce the influence of sparse residual points and poorly defined boundaries representative of real-world scanning conditions, the 5th and 95th percentiles of the projected point coordinates are used for edge estimation rather than the absolute minimum and maximum values. As illustrated in Figure 5(c), these percentile limits define the estimated rectangular boundaries of the projected planar points while reducing the influence of data noise. The resulting rectangular geometry is then validated against the known tag dimensions of 0.60 m × 0.40 m, with a dimensional tolerance of ±0.05 m applied to both the long and short axes. This tolerance was introduced to accommodate variations caused by non-uniform point density, residual measurement noise, imperfect edge definition, unfavourable scan incidence, and local differences in sensor trajectory, while remaining sufficiently restrictive to reject components whose dimensions differ substantially from the designed tag geometry. In addition, the estimated aspect ratio is required to lie at the nominal tag aspect ratio of 1.50 ± 0.10 providing additional tolerance for variations in the estimated boundaries. The aspect-ratio criterion provides an additional geometric constraint to reduce false detections from other planar mine objects that may partially satisfy the dimensional tolerance but do not exhibit the characteristic rectangular proportions of the proposed tag. Components satisfying both the dimensional and aspect-ratio criteria are classified as georeferencing tags.

Following successful validation, the percentile-estimated boundaries are used to determine the tag centre and principal-axis orientation, while the final rectangular geometry is reconstructed using the known nominal dimensions of exactly 0.60 m × 0.40 m. Accordingly, the four corners are positioned at ±0.30 m along the identified long axis and ±0.20 m along the short axis from the estimated tag centre. The transition from the percentile-estimated boundaries to the dimensionally constrained 0.60 m × 0.40 m rectangle is illustrated in Figure 5(c), while Figure 5(d) shows the resulting tag geometry in 3D with its estimated centre and four reconstructed corners. This dimensional constraint prevents residual edge noise or incomplete boundary sampling from propagating into the corner coordinates used for subsequent registration and georeferencing.

Overall, the proposed tag identification procedure, as illustrated in Figure 4-5, provides a robust means of automatically identifying and segmenting the generic georeferencing tags from the surrounding stope point cloud. The sequential use of spatial connectivity, planar fitting, and predefined geometric constraints enables the method to remain tolerant to measurement noise, sparse outliers, imperfectly defined tag boundaries, and variations in point sampling while retaining the characteristic geometry required for reliable tag identification. The resulting tags with their estimated centres and four corners subsequently provide the geometric inputs required for the registration and georeferencing stage of 3D-TARGeT.

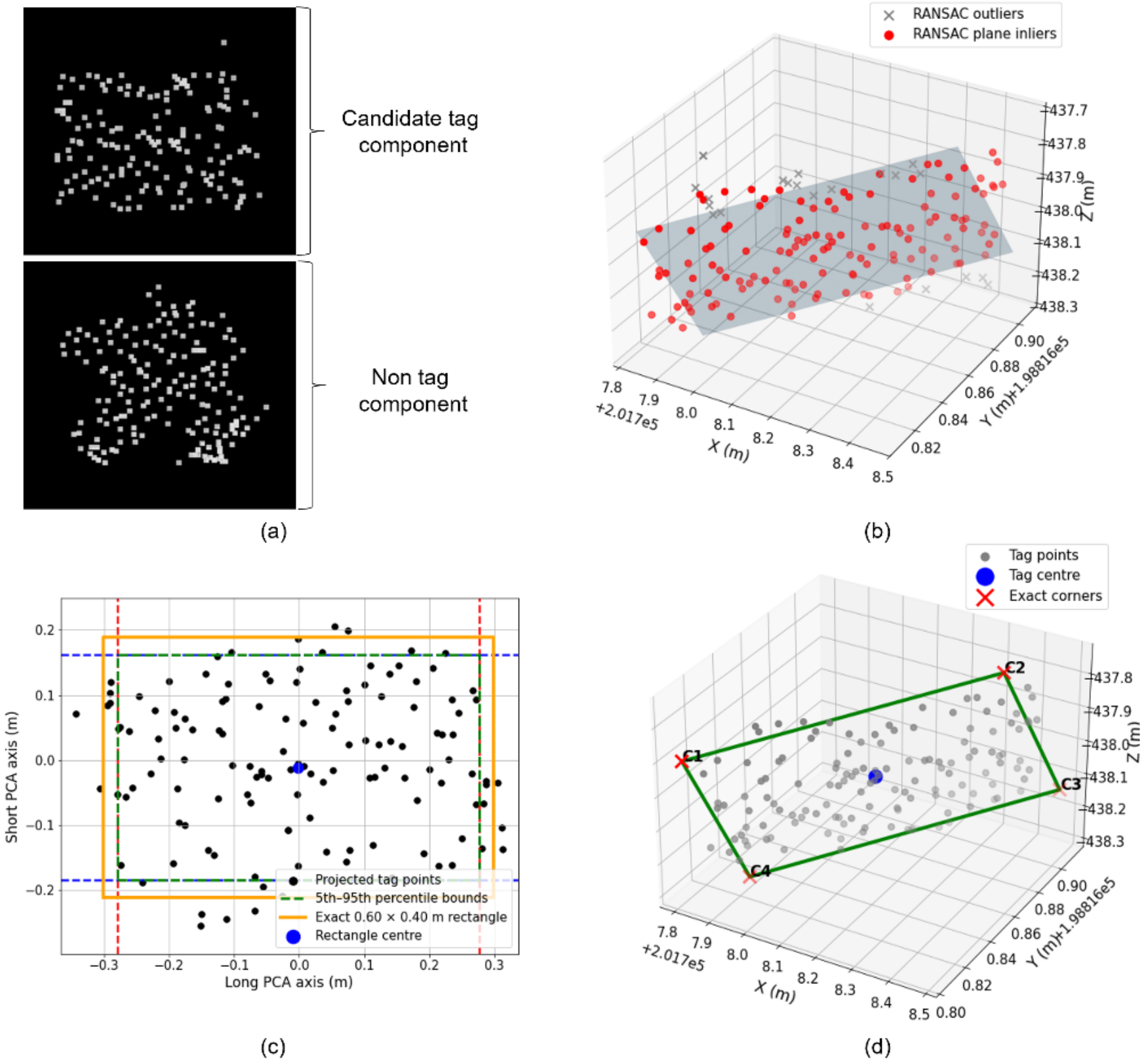

***Figure 5.*** *Sequential identification of the generic georeferencing tag using 3D-TARGeT: (a) representative components isolated by connected-component segmentation, including a candidate tag component with inherent noise and undefined edges and a non-tag component representing a random stray mine feature; (b) RANSAC segmentation of the candidate component showing the fitted plane, planar inliers, and rejected outliers; (c) PCA projection of the planar inliers with 5th–95th percentile edge estimation and reconstruction of the nominal 0.60 m × 0.40 m tag geometry; and (d) final identified tag showing the reconstructed rectangular boundary, estimated centre, and four corners used for subsequent registration and georeferencing.*

### 2.5. Registration and Georeferencing

Following tag identification, the registration and georeferencing stage of 3D-TARGeT uses the estimated centres and corners of the detected tags to determine the rigid transformation between each independently captured stope point cloud and the georeferenced mine reference drives as displayed in Figure 4. The corresponding geometric information of the tags within the reference drives, including their centres and corners, is also required for establishing the subsequent correspondences. If this information is already available from the mine reference dataset, it can be directly used for registration. Alternatively, where only the reference drive geometry containing the tags is available, the mine reference drive shapefiles can first be uniformly downsampled to generate a point-cloud representation. The resulting reference point cloud can then be passed through the tag identification procedure described in Section 2.4 to automatically identify the reference tags and extract their centres and four corners. This provides the reference tag geometry required for matching without requiring manual extraction of individual tag locations.

Since the proposed tags are intentionally generic and non-unique, individual tags do not contain predefined identifiers that directly establish correspondence between the stope scan and the reference dataset. Therefore, the first stage of registration determines the most probable tag correspondences geometrically through triangle-based tag matching. All possible triplets of detected tag centres within the stope scan are enumerated and compared against possible triplets formed by the reference tag centres. The relative inter-tag geometry of each triangle provides a transformation-independent geometric signature, allowing plausible correspondences to be generated without requiring unique tag IDs. For each resulting triangle-matching hypothesis, an initial rigid transformation is estimated using the Kabsch algorithm. The Kabsch algorithm was selected over iterative registration approaches due to its computational efficiency and its closed-form determination of the rigid transformation without requiring iterative convergence or an initial transformation estimate. The Kabsch algorithm determines the rotation and translation that minimise the least-squares positional difference between the corresponding tag centres while preserving the rigid geometry of the point cloud. The resulting rigid transformation can be represented using the 4×4 homogeneous transformation matrix as shown in Equation 2. For each tag centre in the unreferenced stope scan, its corresponding transformed position in the mine reference coordinate system is obtained using Equation 3.

$$T = \begin{bmatrix} a & b & c & t_x \\ d & e & f & t_y \\ g & h & i & t_z \\ 0 & 0 & 0 & 1 \end{bmatrix} \tag{2}$$

*Where T is the rigid transformation matrix, the upper-left 3×3 submatrix represents the rotation and $t_x$, $t_y$ and $t_z$ represent translation along the three coordinate axes.*

$$\begin{bmatrix} x_{ref} \\ y_{ref} \\ z_{ref} \\ 1 \end{bmatrix} = T_{4\times4} \times \begin{bmatrix} x_{org} \\ y_{org} \\ z_{org} \\ 1 \end{bmatrix} \tag{3}$$

*Where a $(x_{org}, y_{org}, z_{org})$ is a point in the unreferenced point cloud, and $(x_{ref}, y_{ref}, z_{ref})$ is the corresponding point in the transformed point cloud.*

For each transformation hypothesis, the root mean square error (RMSE) between the transformed scan tag centres and their corresponding reference tag centres is calculated to quantify the centre-level registration agreement. Rather than immediately selecting the hypothesis producing the minimum mean tag-centre RMSE, all hypotheses with a centre RMSE of ≤0.05 m are retained as candidate transformations. The 0.05 m threshold provides a sufficiently restrictive tolerance relative to the tag dimensions while allowing for small positional uncertainties introduced during tag identification and centre estimation. The inherent spatial variation in tag placement makes it unlikely that multiple correspondence hypotheses will produce similarly low centre RMSE values. Nevertheless, because the proposed tags are intentionally non-unique, a small possibility of geometrically ambiguous correspondence remains. Therefore, rather than relying solely on the minimum tag-centre RMSE, the retained hypotheses are subjected to an additional scan–drive overlap assessment as a secondary validation criterion. This provides an additional safeguard to verify the selected transformation in the unlikely event that more than one plausible tag-matching hypothesis satisfies the centre-RMSE threshold.

To provide an additional level of spatial verification, each retained candidate transformation is subsequently evaluated against the mine reference drives using the scan-drive overlap, reproducing the geometric information conventionally used during manual georeferencing. For each candidate transformation, the corresponding rigid transformation is applied to the stope point cloud and the spatially overlapping regions between the transformed scan and the georeferenced reference drives are identified using a nearest-neighbour distance threshold of 0.10 m. This threshold allows locally corresponding scan and drive surfaces to be identified while limiting the inclusion of spatially separated, non-corresponding points. The RMSE of these overlapping scan-drive points is then calculated, and the candidate transformation producing the minimum overlap RMSE is selected. Evaluating the alignment using the RMSE of the spatially overlapping tunnel regions is important because the extent of tunnel geometry visible within each stope scan can vary substantially across the dataset. Using the number or proportion of overlapping points alone could therefore favour transformation hypotheses with greater spatial coverage, without necessarily indicating better registration. In contrast, the overlap RMSE evaluates the spatial agreement of the corresponding regions and therefore provides a measure of the quality of alignment that is less dependent on the extent of the available tunnel geometry.

Following selection of the most geometrically consistent transformation, the four reconstructed corners of each matched tag are used for corner-based matching and refinement. Unlike the tag centres, which provide one correspondence point per tag, the known rectangular geometry provides four additional spatially distributed correspondence points for each identified tag. The corner correspondences are established between the transformed scan tags and their respective reference tags, and the corner RMSE is calculated to quantify their alignment. The matched corner coordinates are then used to refine the rigid transformation, providing a more spatially constrained estimate of the relative rotation and translation than centre correspondence alone. The use of the dimensionally constrained 0.60 m × 0.40 m tag geometry established in Section 2.4 ensures that residual variations in scanned tag boundaries do not directly propagate into this refinement stage.

The refined rigid transformation is finally applied to the complete stope point cloud using the transformation represented in Equations 2 and 3, thereby transforming the scan from its local scanner reference frame into the established mine coordinate system. The same procedure is independently applied to each multi-temporal stope scan, producing a set of consistently registered and georeferenced point clouds within the common mine reference framework. The resulting georeferenced multi-temporal dataset can subsequently be directly compared across successive stope stages and integrated with existing mine spatial information for downstream geological and geotechnical analysis.

### 2.6. Validation Techniques

The accuracy of the proposed 3D-TARGeT framework was evaluated against manually georeferenced stope point clouds used as the reference dataset. For each multi-temporal scan, manual georeferencing was performed by visually aligning the tunnel and access-drive regions captured within the stope point cloud with the corresponding georeferenced mine reference drives. The alignment was achieved through iterative manual translation and rotation until the common tunnel geometries were considered to be spatially aligned. The resulting manually georeferenced point clouds were subsequently used as the reference for evaluating the automated georeferencing results. Although manual alignment may be subject to minor human interpretation error, it represents the conventional georeferencing procedure applied to the investigated dataset and was therefore adopted as the ground-truth reference for evaluating the accuracy of the automated georeferencing results.

To benchmark the proposed framework against conventional point-cloud registration approaches, the output of 3D-TARGeT was further compared with two widely used rigid registration techniques, namely Iterative Closest Point (ICP) and Normal Distributions Transform (NDT). Both methods were implemented independently of the proposed tag-based workflow to establish correspondence and estimate the rigid transformation between each multi-temporal stope point cloud and the georeferenced mine reference drives, without using the identified georeferencing tags or any tag-derived initial transformation. ICP iteratively establishes closest-point correspondences between the target and reference point clouds and updates the rigid transformation to minimise the spatial discrepancy between the corresponding points. Despite its widespread use and computational simplicity, ICP is inherently dependent on sufficient common geometry and is sensitive to the initial relative alignment of the datasets, with inadequate initialisation potentially resulting in convergence to an incorrect local solution. NDT provides a fundamentally different registration strategy by discretising the reference point cloud into a regular voxel grid and statistically representing the spatial distribution of points within each populated voxel, typically using a multivariate normal distribution. Registration is then performed by optimising the transformation of the target point cloud relative to this probabilistic representation of the reference geometry rather than explicitly establishing individual nearest-point correspondences. A voxel size of 0.5 m was adopted for the NDT implementation in this study, providing a balance between retaining the geometric characteristics of the tunnel surfaces and avoiding an excessively fine voxel representation that would increase computational demand and reduce the number of observations available for estimating the local distributions. Since NDT performance is influenced by the selected voxel resolution and the amount of common geometric structure between the datasets, it provides a complementary distribution-based benchmark to the correspondence-based ICP approach.

### 2.7. Structure Mapping

Routine assessment of geological structures and discontinuity planes in underground excavations such as stopes is an integral component of geotechnical assessment for evaluating rock mass stability and supporting the operational efficiency and safety of underground mines. Discontinuities represent planes of weakness within the geological rock mass, and their characterisation provides important information for ground stability assessment, mine planning, and design. Structure mapping is a well-established field, with extensive research conducted on automated and semi-automated discontinuity plane extraction from 3D point cloud data. It is one of the most important downstream geotechnical analyses routinely undertaken by geological and geotechnical engineers at mine sites. Routine assessment of geological structures and discontinuity planes in underground excavations such as stopes is an integral component of geotechnical assessment for evaluating rock mass stability and supporting the operational efficiency and safety of underground mines. Discontinuities represent planes of weakness within the geological rock mass, and their characterisation provides important information for ground stability assessment, mine planning, and design. Structure mapping is a well-established field, with extensive research conducted on automated and semi-automated discontinuity plane extraction from remotely sensed three-dimensional data and it represents one of the important downstream geotechnical analyses routinely undertaken by geological and geotechnical engineers at mine sites (Battulwar et al., 2021; Chen et al., 2023; Daghigh et al., 2022; Pu et al., 2025). Importantly, the practical application of structure mapping to stope point clouds is dependent on accurate georeferencing, as the spatial locations and orientations of the identified structural features must be represented within the mine reference coordinate system.

Structure mapping of multi-temporal stopes presents an additional challenge, as successive stope datasets need to be accurately registered and georeferenced to the same reference frame before structural features can be consistently compared and tracked over time. However, the time-consuming nature of conventional manual registration and georeferencing can restrict the routine processing of large volumes of stope point-cloud data and contribute to substantial data backlogs, leaving potentially valuable datasets unprocessed. The ability to consistently georeference successive stope scans can therefore provide a basis for time-series tracking of structural planes throughout stope development. Such temporal analysis can provide additional geotechnical value by facilitating the comparison and correlation of the structural fabric exposed at successive stages, with potential applications in ground-condition assessment, future stope and blast design, and mine planning.

The generic rectangular georeferencing tags and the proposed 3D-TARGeT methodology provide a framework for accurately registering and georeferencing both individual and multi-temporal stope point clouds within the mine reference coordinate system. This provides the spatial basis required for conventional structure mapping and, importantly, enables the resulting structural information from successive scans to be represented and compared within a common reference framework. In this study, a structure-mapping framework was implemented using well-established automated clustering-based structure-mapping techniques as its basis (Patra, Ranasinghe, et al., 2025b; Patra et al., 2026; Riquelme et al., 2018; Singh et al., 2021), accompanied by manual supervision for refinement and verification, and applied as a downstream application of 3D-TARGeT to demonstrate this capability. No new structure-mapping methodology is proposed as part of this study. Rather, established techniques are used to demonstrate the practical advantage provided by automated georeferencing for the analysis of multi-temporal stope datasets. The mapped discontinuity orientations from the successive scans were subsequently represented within a common stereographic reference framework to facilitate time-series comparison and tracking of the identified structural planes.

## 3. Results and Discussions

### 3.1. Tag Identification

The tag identification procedure described in Section 2.4 demonstrated robust performance under the challenging underground mine conditions and inherent point-cloud noise present within the multi-temporal stope scans. All generic rectangular georeferencing tags simulated and incorporated across the investigated scans were successfully recognised by the 3D-TARGeT tag identification stage, achieving an identification accuracy of 100% without any false-positive detections. In contrast, the use of CC segmentation alone produced several non-tag candidate components due to the presence of spatially isolated mine objects and features. Moreover, even the components containing the actual tags frequently included residual noise, suspension elements, and adjacent stray features, preventing CC segmentation alone from reliably defining the tag geometry. The subsequent RANSAC plane segmentation and PCA-based geometric validation incorporated within 3D-TARGeT effectively refined these candidate components, isolated the planar tag surfaces, and reconstructed the nominal tag geometry despite imperfect and noisy boundaries. As illustrated in Figure 6, the framework consistently identified the generic tags within the complex stope environments and accurately reconstructed their 0.60 m × 0.40 m boundaries, centres, and four corners, providing the geometric information required for the subsequent registration and georeferencing stage.

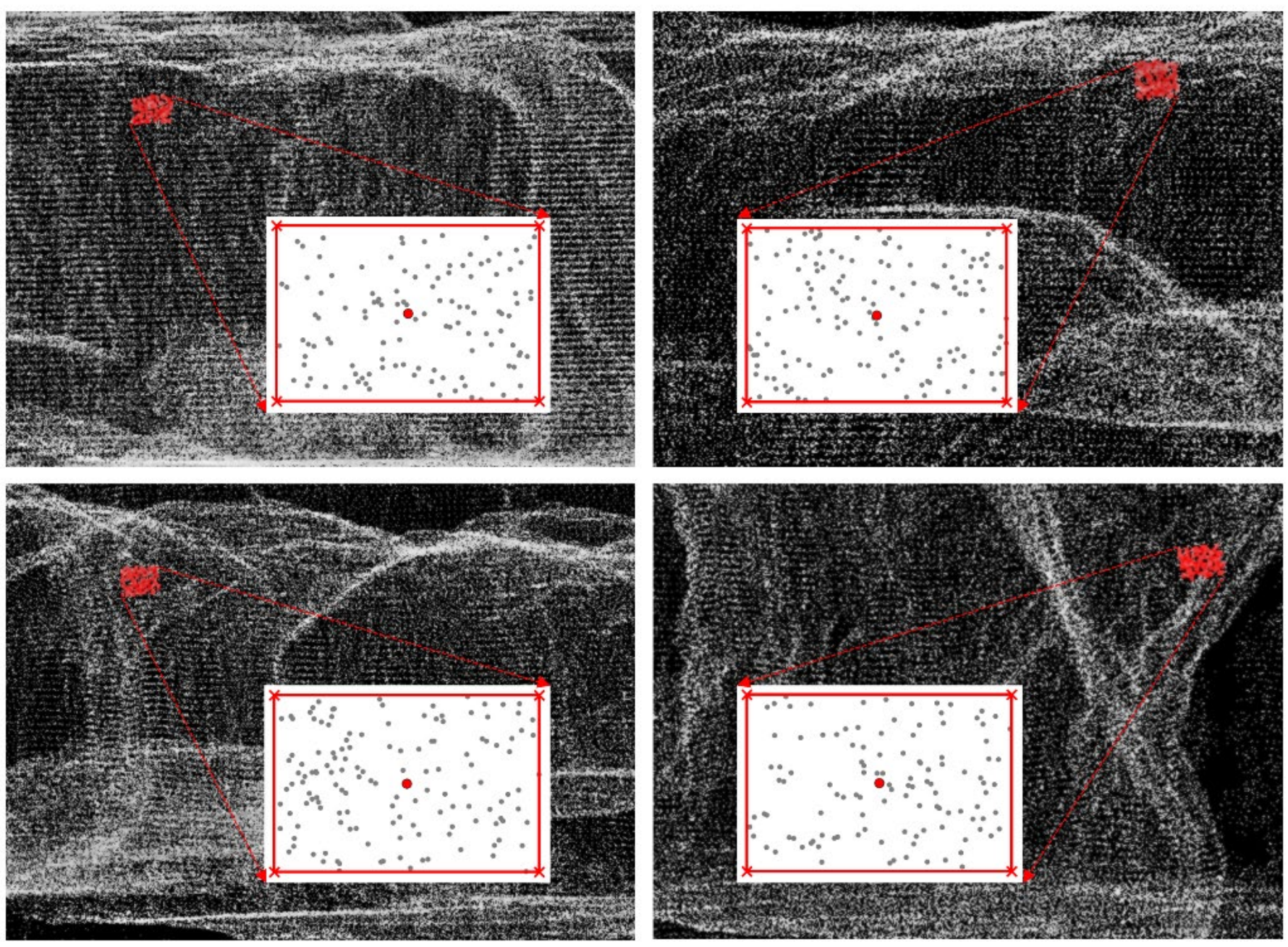

***Figure 6.*** *Identified candidate generic rectangular georeferencing tags within the stope point clouds, with enlarged views showing the identified tag points, with the reconstructed boundaries, centres, and four corners.*

### 3.2. Registration and Georeferencing

Following successful tag identification, the detected tags were matched with their corresponding reference tag centres and corners in the reference mine drives to estimate and refine the transformation required for registration and georeferencing. Figure 7 presents the tag-matching and refined georeferencing results for the four multi-temporal stope scans. Reference tag centres are shown as green circles, while the corresponding transformed scan tag centres are represented by red crosses, with the residual centre distances indicated for each matched tag. The final computed transformation matrix, number of detected and successfully matched tags, and refined tag RMSE are also presented for each scan. For all four scans, all detected tags were successfully matched to their corresponding tags in the reference drive, despite the number of available tags varying considerably from 3 to 18 across the multi-temporal scans. The residual centre RMSE error remained consistently in the low single-digit-centimetre level across all scans. Importantly, no significant variation in the registration RMSE was observed with increasing or decreasing numbers of available tags, indicating that the proposed matching and refinement procedure maintained consistent registration accuracy across scans with substantially different numbers of visible tags. The resulting refined transformation matrix for each scan was subsequently applied to the respective complete stope point cloud to obtain the final georeferenced datasets.

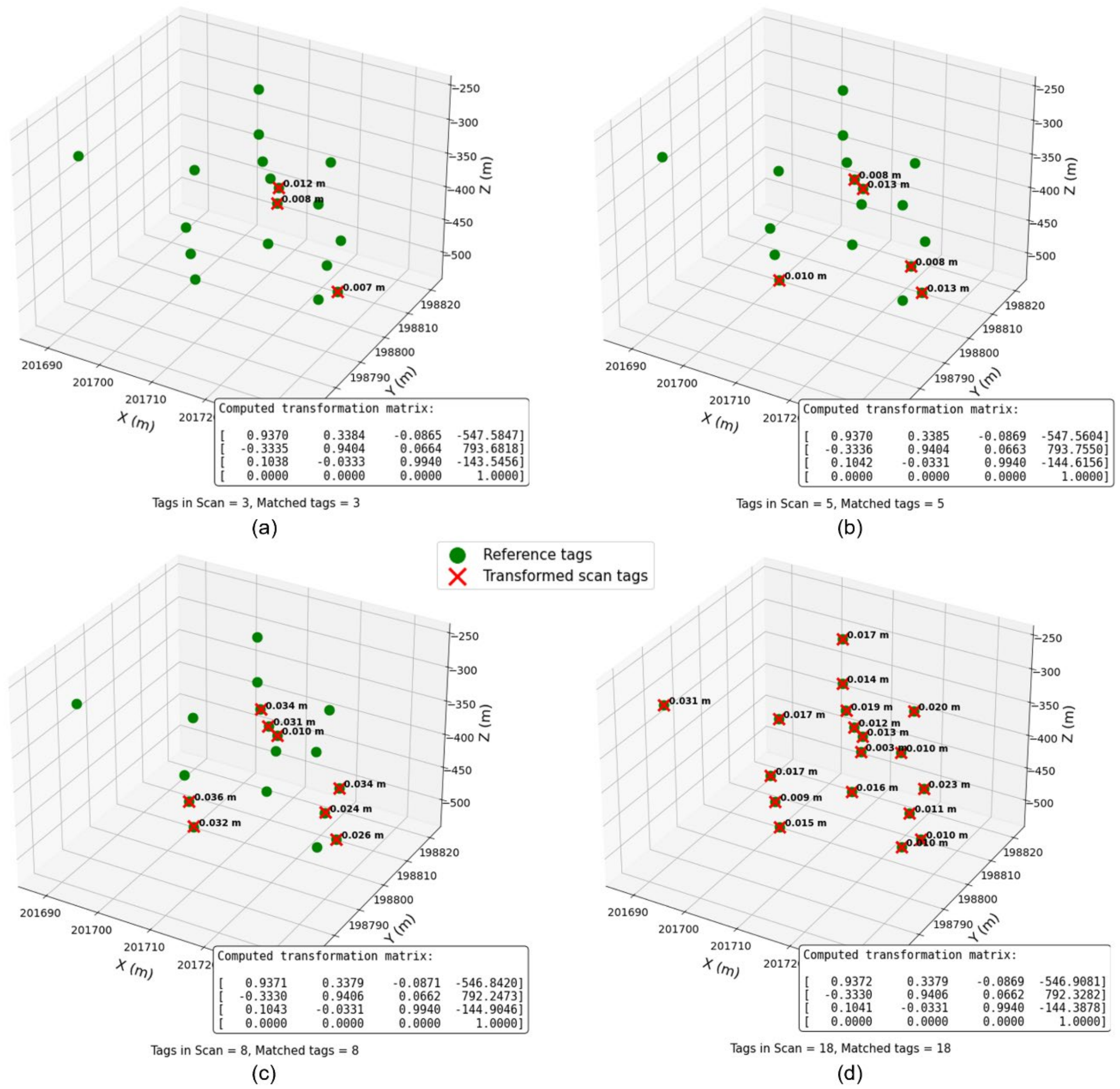

***Figure 7.*** *Tag-matching results leading to the refined registration and georeferencing of the four multi-temporal stope point clouds: (a) Scan 1, (b) Scan 2, (c) Scan 3, and (d) Scan 4, showing the refined centre and corner RMSE and the final transformation matrix for each scan.*

### 3.3. 3D-TARGeT Validation

The geometric agreement between the automatically and manually georeferenced point clouds was quantified using cloud-to-cloud (C2C) distance and the corresponding root mean square error (RMSE), as presented in Figure 8. For each point within the evaluated 3D-TARGeT result, the Euclidean distance to its nearest neighbouring point in the manually georeferenced reference cloud was calculated. The resulting point-wise C2C distances therefore provide a direct measure of the spatial separation between the automatically and manually aligned point-cloud surfaces. Rather than relying solely on a single aggregate error value, the distribution of C2C distances was examined to characterise both the central tendency and spatial variability of the georeferencing error and to identify any substantial local discrepancies within the registered geometry. The overall RMSE error was additionally calculated as an overall measure of registration accuracy. As the squared-error term places greater emphasis on larger discrepancies, RMSE complements the C2C distance distribution by providing a single metric that is more sensitive to comparatively large local misalignments.

Figure 8a shows the C2C distance distributions obtained for the four multi-temporal stope scans following automatic georeferencing using 3D-TARGeT, with the manually georeferenced point clouds used as the reference. A consistently narrow distribution was observed across all four scans, with the interquartile range remaining below approximately 0.02 m irrespective of the scan. The median C2C distance also remained below 0.03 m for all scans, indicating that the typical spatial discrepancy between the automatically and manually georeferenced surfaces was limited to low single-digit centimetres. Importantly, none of the scans exhibited a substantial increase in either the median or spread of the C2C distances, suggesting an absence of pronounced scan-specific registration discrepancies. A modest decreasing trend in median C2C distance can nevertheless be observed from Scan 1 to Scan 4, coinciding with an increase in the number of available tags from 3 to 18. The additional tags provide greater spatial redundancy for estimating and refining the transformation and may therefore contribute to this improvement. However, the difference in median C2C distance between Scan 1 and Scan 4 is only approximately 0.01 m, despite the six-fold increase in the number of available tags. This relatively small change is particularly significant because it indicates that high registration accuracy was already achieved using only three tags, while the availability of additional tags primarily increased the redundancy of the transformation rather than being essential for obtaining an accurate solution.

A consistent pattern is observed in the RMSE results presented in Figure 8b. The RMSE remained below 0.03 m across all four scans and decreased gradually from approximately 0.028 m for Scan 1 to approximately 0.014 m for Scan 4. The agreement between the low median C2C distances, narrow C2C distributions and low RMSE values provides complementary evidence of the quality of the resulting georeferencing. In particular, the low RMSE indicates that the C2C distributions are not dominated by comparatively large local errors that could otherwise be obscured by the median alone. Collectively, these results demonstrate consistent centimetre-level agreement between the 3D-TARGeT outputs and manually established reference across all four multi-temporal scans, despite differences in scan geometry and the number of tags available for registration.

The limited variation in registration accuracy with tag number is also important from a practical implementation perspective. Although the results indicate a gradual improvement as additional tags become available, the comparatively small difference between the scans demonstrates that the proposed registration procedure is not strongly dependent on a large number of visible tags once the minimum geometric constraints required for rigid transformation are satisfied. This characteristic is particularly relevant for multi-temporal stope monitoring, where the number of tags captured within individual scans may vary because of differences in scan coverage, excavation progression, visibility and accessibility. In general, underground stopes are accessible from more than three access points across different levels, allowing the minimum required number of spatially distributed tags to be installed at separate access locations without requiring multiple tags at the same access point. This keeps the physical implementation comparatively simple while still providing sufficient geometric constraints for the proposed registration procedure. Where a larger number of tags are visible, they provide additional spatial redundancy and further constrain the transformation. Overall, the consistency of both the C2C distance distributions and RMSE across the four scans demonstrates the accuracy and robustness of 3D-TARGeT under varying stope geometries and tag availability.

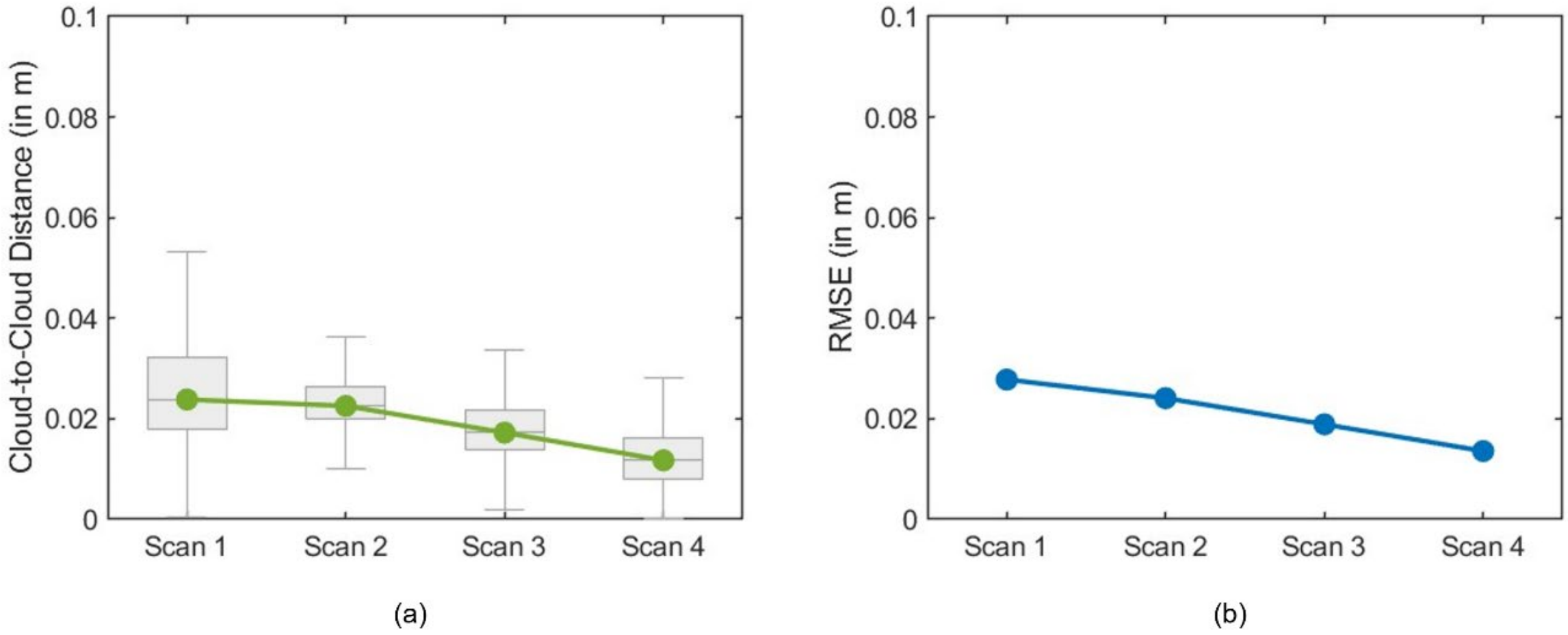

***Figure 8.*** *Accuracy assessment of the 3D-TARGeT georeferencing results across the four multi-temporal stope scans: (a) cloud-to-cloud (C2C) distance distributions and median C2C distances, and (b) corresponding C2C RMSE for each scan.*

To further evaluate the performance of 3D-TARGeT, its registration and georeferencing results were benchmarked against two widely used automatic point-cloud registration techniques, ICP and NDT.

Both methods determine a rigid transformation by exploiting geometric similarity between a target point cloud and a reference dataset, although they achieve this through different registration principles as described in Section 2.6. In the present application, the objective is specifically georeferencing rather than co-registering successive stope scans. Accordingly, the unreferenced stope point cloud was treated as the target and registered directly against the georeferenced mine reference drives, such that the estimated transformation would place the stope scan within the established mine coordinate system.

Scan 4 was selected for this comparative analysis because it contains the greatest extent of common tunnel and access-drive geometry with the mine reference drives among the four multi-temporal scans. This provides ICP and NDT with the most favourable case for conventional geometry-based registration, as both techniques depend on the presence of sufficient common geometric information between the target and reference datasets. The comparison therefore does not intentionally disadvantage the conventional approaches by selecting a scan with particularly limited overlap. Instead, Scan 4 provides the greatest opportunity for ICP and NDT to determine an appropriate transformation from the available common geometry. For consistency, ICP, NDT and 3D-TARGeT were applied to the same unreferenced Scan 4 point cloud and corresponding georeferenced reference-drive dataset. The resulting point clouds were subsequently evaluated against the same manually georeferenced Scan 4 reference using the C2C distance and RMSE measures described previously. The comparison consequently assesses both the positional accuracy achieved by each method and its ability to determine the required rigid transformation under the geometric conditions characteristic of stope-to-drive georeferencing.

The resulting differences between the three approaches are visually apparent in Figure 9. For ICP Figure 9a, substantial separation remains between the automatically georeferenced point cloud and the manually georeferenced point cloud, demonstrating that the algorithm converged to an incorrect spatial alignment. As ICP generally requires an approximate initial alignment, a rough alignment between the unreferenced stope scan and the mine reference drives was provided prior to registration to give the method a favourable starting condition. Despite this initialisation, the resulting transformation still exhibited substantial misalignment, since nearest-point correspondences could be established between non-corresponding or geometrically similar surfaces, potentially directing the iterative optimisation towards an incorrect local solution. NDT Figure 9b produces a comparatively improved alignment, with portions of the common geometry brought closer together. Unlike ICP, NDT does not rely on explicit point-to-point correspondences, but instead represents the reference geometry through voxel-level probability distributions and optimises the alignment relative to these local spatial distributions. This distribution-based representation appears to provide greater tolerance to the limited common geometry between the target and reference drives, resulting in comparatively better alignment than ICP. However, the limited overlap remains insufficient for accurate global georeferencing, with a substantial positional discrepancy still evident between the NDT-transformed scan and the manually established georeferenced position. In contrast, the 3D-TARGeT result in Figure 9c visually overlaps closely with the manually georeferenced point cloud, to the extent that the two datasets are largely coincident at the scale of the complete stope.

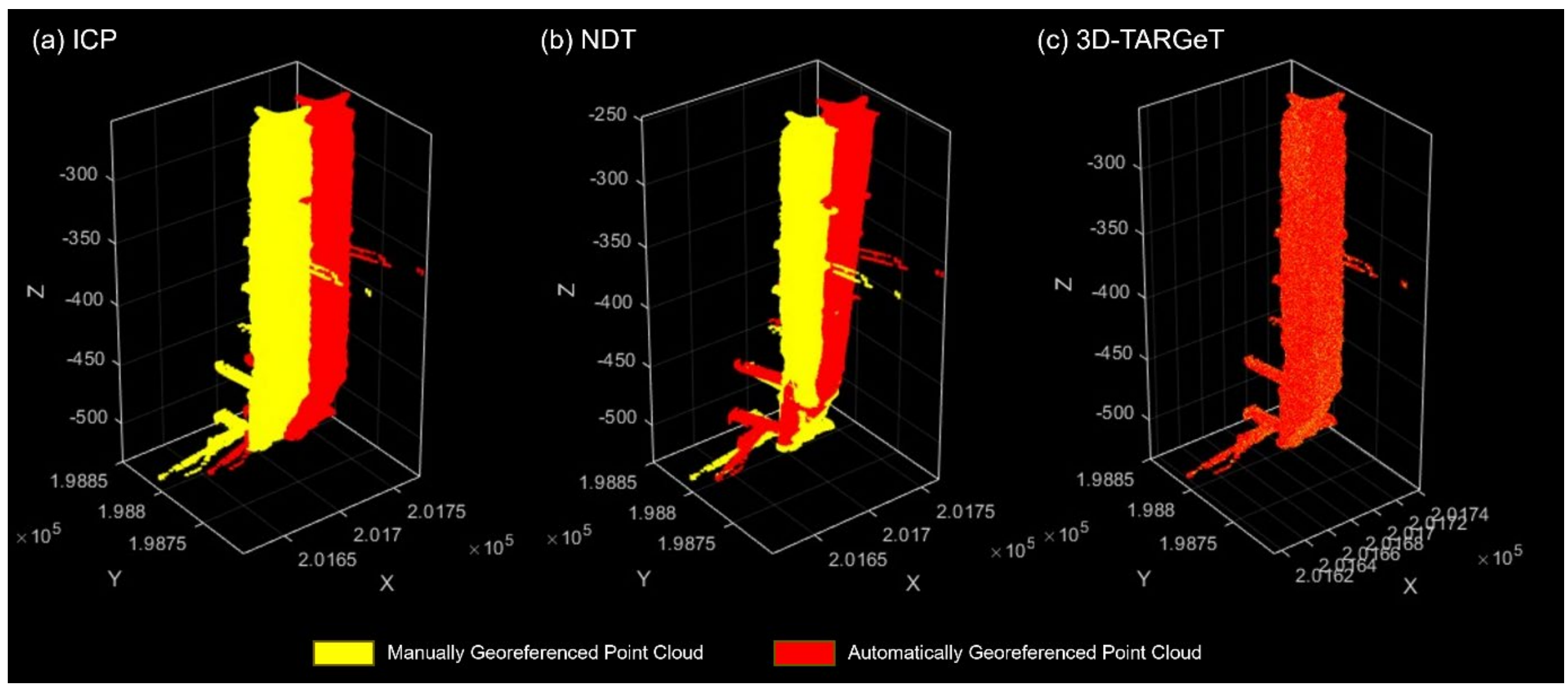

***Figure 9.*** *Visual comparative analysis of the registration and georeferencing results obtained using (a) ICP, (b) NDT, and (c) the proposed 3D-TARGeT framework on Scan 4, showing the spatial alignment between the manually georeferenced point cloud and the automatically georeferenced point cloud.*

The quantitative results presented in Table 1 further support the differences observed visually in Figure 9. Among the two conventional registration techniques, NDT demonstrated a clear improvement over ICP, reducing the median C2C distance by approximately 64% and the RMSE by approximately 53%. Despite this improvement, both conventional approaches retained errors at the metre scale, whereas the proposed 3D-TARGeT framework achieved centimetre-level agreement with the manually georeferenced reference, with a median C2C distance of approximately 1.2 cm and an RMSE of approximately 1.4 cm. This corresponds to a greater than 99.9% reduction in both median C2C distance and RMSE compared with ICP, and approximately 99.8% reduction compared with NDT. The comparison therefore demonstrates a substantial improvement in positional accuracy using the proposed tag-based georeferencing approach compared with the conventional geometry-based registration techniques.

The substantial difference can principally be attributed to how the correspondence problem is addressed. ICP and NDT attempt to recover the transformation directly from the geometric relationship between the two point-cloud datasets. This becomes challenging for the present application because a stope scan contains extensive excavation geometry that is absent from the reference-drive dataset, while only the comparatively limited access-drive regions provide common geometry. Repeated or locally similar tunnel surfaces can further provide ambiguous geometric information, particularly when the datasets begin with substantial translational and rotational offsets. 3D-TARGeT instead introduces explicit geometric constraints through the identified tags. The tag-matching stage establishes correspondence between spatially distributed locations in the unreferenced scan and the reference drives before the complete stope point cloud is transformed. Consequently, the registration is not dependent on discovering the correct global alignment solely from partially overlapping excavation surfaces. The subsequent centre-based transformation, corner-based refinement, and overlap-based verification further constrain the solution, accounting for the close agreement observed with the manually georeferenced result.

A notable difference is also evident in computational performance. All experiments were conducted using Python 3.10.9 on a 64-bit system equipped with an Intel Xeon W-2245 3.9 GHz CPU and 128 GB DDR4 memory. ICP required 69.65 s, while 3D-TARGeT required 73.59 s, demonstrating comparable computational times despite the substantially higher positional accuracy achieved by 3D-TARGeT. In contrast, NDT required approximately 70 min, making it considerably more computationally intensive. This large computational requirement arises from the voxel-based probabilistic formulation of NDT, where the reference point cloud must first be discretised into voxels and local probability distributions estimated, followed by iterative evaluation and optimisation of the transformed target points against these distributions. For the large 3D stope and reference-drive datasets considered in this study, these repeated voxel-level probabilistic evaluations resulted in a substantially greater computational burden. ICP is comparatively inexpensive because of its computationally efficient local-registration formulation, where correspondence search can be accelerated through spatial indexing and nearest-neighbour lookup, such as KD tree-based search, followed by efficient rigid-transformation estimation and rapid local convergence when an appropriate initial alignment is provided. Similarly, 3D-TARGeT maintains a relatively low computational cost by avoiding repeated optimisation over the complete stope geometry. Instead, the registration is primarily performed using the comparatively small number of identified tag centres and corners, with efficient geometric matching and closed-form rigid-transformation estimation used to determine the transformation before it is applied once to the complete point cloud. Consequently, 3D-TARGeT achieved centimetre-level georeferencing accuracy while maintaining a computational requirement comparable to ICP and substantially lower than NDT.

***Table 1.*** *Statistical comparative analysis of the different georeferencing and registration techniques on Scan 4 against the manually georeferenced point cloud.*

| **Georeferencing and Registration Method** | **Median C2C Distance (in m)** | **RMSE Error (in m)** | **Processing Time (in seconds)** |
|---|---|---|---|
| ICP | 16.6238 | 19.5641 | 69.65 |
| NDT | 6.0524 | 9.1468 | 4207.09 |
| 3D-TARGeT | 0.0117 | 0.0136 | 73.59 |

Overall, the benchmark highlights an important distinction between conventional point-cloud registration and the proposed georeferencing framework. ICP and NDT remain effective registration approaches when sufficient common geometry and appropriate initial conditions are available. However, the results demonstrate their limitations when directly georeferencing a large stope scan against mine reference drives with limited common geometry and without a prior transformation. Even under the comparatively favourable conditions provided by Scan 4, neither conventional method recovered the manually established georeferenced position with sufficient accuracy. In contrast, 3D-TARGeT achieved centimetre-level agreement while maintaining a processing time comparable to ICP. Combined with the consistent performance across the four multi-temporal scans demonstrated in Figure 8, these results indicate that explicitly establishing correspondence through generic georeferencing tags provides a more reliable basis for automated registration and georeferencing in this application than relying solely on naturally occurring common excavation geometry. The resulting accurately georeferenced multi-temporal stope point clouds can subsequently be integrated within a common mine coordinate framework, providing spatially consistent datasets for downstream geotechnical analyses, mine planning and engineering assessments, as well as the development of automated spatial monitoring and analysis applications.

More importantly, 3D-TARGeT removes the need for the time-intensive manual georeferencing currently required to spatially align individual stope scans with the mine reference framework. By enabling multi-temporal stope point clouds to be georeferenced automatically, efficiently and robustly within a common mine coordinate system, the framework can facilitate the processing of larger volumes of spatial data that may otherwise be constrained by the manual georeferencing requirement. This increases the availability of consistently georeferenced multi-temporal datasets for downstream geotechnical analysis, time-series analysis, mine planning and engineering assessments, while also providing a foundation for further automated spatial monitoring and analysis applications within underground mining operations.

### 3.4. Structure Mapping

Figure 10 presents the time-series structure mapping results obtained from the four stope scans following their registration and georeferencing using 3D-TARGeT. Discontinuity planes are geologically represented by their dip angle (DA) and dip direction (DD), where dip angle describes the inclination of a discontinuity plane from the horizontal and dip direction represents the azimuth of its downward inclination, measured clockwise from true north. Consequently, the geological orientation assigned to a discontinuity has meaningful spatial context only when the point cloud is accurately georeferenced within the mine reference coordinate system. The individual stereonets in Figure 10 show the discontinuity orientations mapped from each of the four temporal scans within this common reference frame. To demonstrate the potential for tracking the structural fabric through successive stages of stope development, the discontinuity planes identified in Scan 1 were considered the initial population, comprising 74 planes. For each subsequent scan, a plane was considered newly identified only when no previously established plane occurred within a tolerance of ±5° in both dip angle and dip direction. This tolerance was adopted to avoid interpreting small orientation variations arising from local surface variability, point-cloud characteristics, or structure-mapping uncertainty as geologically distinct planes. Accordingly, the analysis provides an overall estimation of the emergence and persistence of the structural fabric rather than an exact one-to-one tracking of individual physical discontinuity surfaces. Using this criterion, 27, 25, and 4 newly identified planes were observed from Scans 2, 3, and 4, respectively, in addition to the initial 74 planes identified in Scan 1. A parula sequential colour map was used in the combined stereonet to distinguish the temporal origin of the mapped structures, with each colour corresponding to the scan in which a discontinuity plane was first identified, as indicated in the legend.

The individual and combined stereonets in Figure 10 further demonstrate that a substantial proportion of the structural fabric observed during the initial stages remains represented throughout the multi-temporal sequence, while additional orientations become exposed as the stope progresses. This consistency indicates that meaningful estimates and interpretations of the geological structural fabric can be obtained from different stages of stope development, rather than being restricted to a single scan or excavation stage. At the same time, newly exposed discontinuities can supplement the existing structural information as additional rock surfaces become available during excavation. This application highlights the potential geotechnical benefits of 3D-TARGeT for one of the important downstream analysis tasks routinely undertaken in underground mines. By automatically registering and georeferencing successive stope point clouds within a common mine reference frame, 3D-TARGeT enables discontinuity orientations obtained at different stages to be directly represented, compared, and tracked within the same geological reference system. The results therefore demonstrate how the proposed georeferencing framework can facilitate the use of multi-temporal stope datasets for routine downstream structure mapping and provide a progressively enriched representation of the structural fabric for subsequent geological and geotechnical assessment.

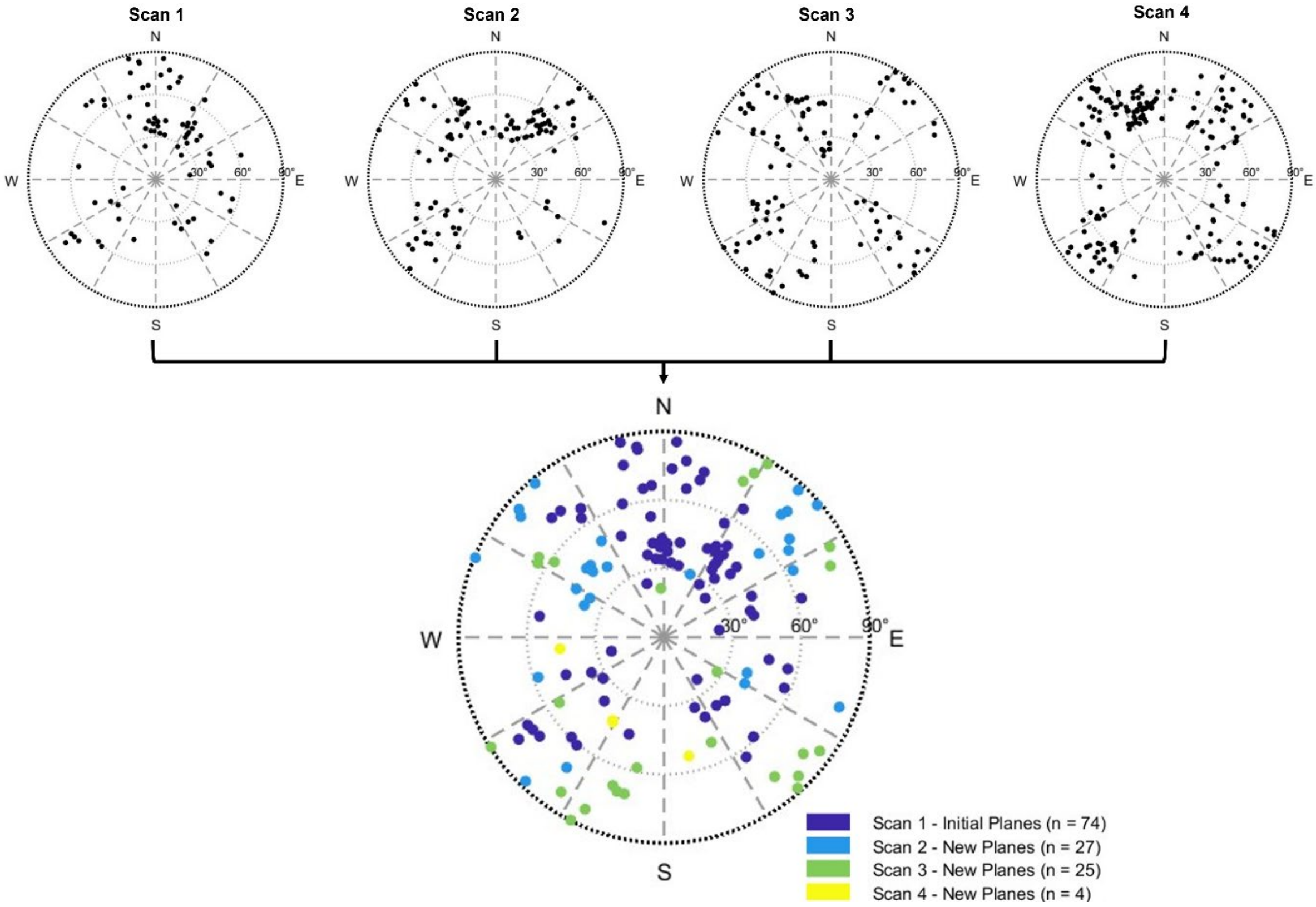

***Figure 10.*** *Time-series structure across the four georeferenced stope scans. Individual stereonets show the discontinuity orientations identified in each scan, while the combined stereonet presents the initial planes from Scan 1 and newly emerging planes in Scans 2-4.*

## 4. Future Research Directions

The present study establishes the feasibility of 3D-TARGeT using simulated tags, with efforts made to reproduce realistic underground scanning conditions, including representative tag geometry, point density, scanning range, and measurement noise. Given the promising registration and georeferencing performance demonstrated in this study, the next stage of research should focus on implementing physical tags across operational underground mines and evaluating the framework under routine production conditions. Such field-scale implementation would enable assessment of tag durability, placement and visibility requirements, performance under varying scanning conditions, and the practical benefits of 3D-TARGeT in reducing manual georeferencing effort and facilitating the processing of larger volumes of routinely acquired point-cloud data.

The ability to repeatedly register spatial datasets within a consistent mine reference frame also provides opportunities beyond the multi-temporal stope application demonstrated in this study. Future research could investigate repeated co-registration of the same excavation regions for automated three-dimensional change detection and deformation monitoring. Potential applications include monitoring roadway and tunnel convergence or subsidence, floor heave, progressive fracturing, and other spatial changes associated with ground behaviour. Although 3D-TARGeT was developed and demonstrated for multi-temporal stope point clouds, the generic nature of the tag-based reference framework also allows its application to other underground excavations where reliable spatial correspondence is required. Permanent or semi-permanent randomly located tags could potentially establish local reference networks along drives, intersections, drawpoints, ore passes, chambers, and other underground infrastructure. This could support repeated spatial monitoring without requiring extensive common natural geometry between successive scans, particularly in environments undergoing substantial geometric change.

Another potential application is automated excavation and blast reconciliation. Consistently georeferenced post-blast point clouds could be directly compared with mine design geometry and previous excavation states to quantify overbreak, underbreak, excavation volumes, and changes associated with successive blasts. When combined with blast design and geological information within the mine reference frame, such datasets could support investigation of relationships between blast performance, excavation geometry, and structural conditions. The same spatial framework could potentially support reconciliation of other mine infrastructure and spatial information, including installed ground support, development profiles, and as-built excavation geometry against their respective designs.

The framework could also facilitate the integration of spatial datasets acquired using different sensing platforms. Underground mines increasingly employ terrestrial laser scanners, mobile laser scanning systems, UAV-mounted LiDAR, and other three-dimensional sensing platforms, which may produce datasets with different local coordinate systems and acquisition characteristics. Persistent georeferencing targets could provide common spatial references through which compatible datasets acquired at different times or using different platforms are brought into the same mine coordinate system. This could enable more systematic integration of spatial information across different stages of mine development and monitoring activities.

A further research direction is the integration of the proposed tags as ground control targets (GCTs) within SLAM-based underground mapping systems. In the present framework, the tags provide spatial references for registration and georeferencing after point-cloud acquisition and map generation. Future development could investigate their direct incorporation into the SLAM process as recognisable spatial constraints, enabling known tag locations to assist with trajectory and map estimation during data acquisition. Beyond passive geometric tags, active or information-embedded GCTs could also be investigated, where individual targets contain directly identifiable information linked to known mine coordinates. Such an approach could potentially eliminate the subsequent tag-matching and registration stages by establishing correspondence with the mine reference frame during map construction itself. These developments could extend 3D-TARGeT from a post-processing georeferencing framework towards an integrated spatial referencing system for automated underground mapping and multi-temporal geotechnical monitoring.

## 5. Conclusions

This study proposed the 3D Tag-based Automated Registration and Georeferencing Technique (3D-TARGeT) to automate the registration and georeferencing of multi-temporal stope point clouds within GNSS-denied underground mine environments. The framework uses low-cost, generic, non-unique rectangular georeferencing tags to establish spatial correspondence between locally referenced stope scans and the established mine reference coordinate system. By combining automated tag identification, geometric tag matching, rigid transformation estimation, spatial verification, and refinement, 3D-TARGeT provides an alternative to the repeated manual alignment currently required to prepare stope point clouds for downstream applications. Across four multi-temporal stope scans, 3D-TARGeT achieved consistent centimetre-level agreement with manually georeferenced reference datasets, with median cloud-to-cloud distances and RMSE remaining below 0.03 m despite the number of available tags varying largely between the scans. Benchmarking further demonstrated substantially higher positional accuracy than ICP and NDT, while maintaining a processing time comparable to ICP and considerably lower than NDT.

The resulting consistently georeferenced multi-temporal datasets provide a spatial basis for downstream geological and geotechnical analyses and applications, as briefly demonstrated through time-series structure mapping, where structural information from successive excavation stages could be represented and compared within a common mine reference framework. As a proof-of-concept study, the proposed tags were simulated and incorporated into acquired stope datasets, with representative tag geometry, point density, and scanner characteristics considered to reproduce practical scanning conditions as closely as possible. The results demonstrate the potential of 3D-TARGeT to streamline georeferencing workflows and facilitate greater utilisation of routinely acquired underground point-cloud data. Future implementation using physical tags in operational mines will enable further evaluation of the framework under real-world conditions and its extension to broader applications including change detection, deformation monitoring, and excavation reconciliation.

**Acknowledgements**

Dibyayan Patra acknowledges the financial support provided by the University International Postgraduate Award (UIPA) from the University of New South Wales for this research. The authors also acknowledge the facilities and resources provided by the Laboratory for Imaging of the Mine Environment at the University of New South Wales, Sydney, for this research.

**Author contributions**

CRediT: **Dibyayan Patra:** Writing – original draft, Data curation, Investigation, Conceptualisation, Methodology, Software, Formal analysis, Visualisation, Validation. **Simit Raval:** Writing – review and editing, Project administration, Data curation, Resources, Supervision. **Pasindu Ranasinghe:** Writing – review and editing, Formal Analysis, Validation. **Bikram Banerjee:** Writing – review and editing, Validation, Supervision. **Ismet Canbulat:** Writing – review and editing, Supervision.

**Declaration of generative AI and AI-assisted technologies in the writing process**

During the preparation of this work, the authors used OpenAI's ChatGPT (version GPT-5.6) to improve the language and readability of the manuscript. After using this tool, the authors reviewed and edited the content as needed and take full responsibility for the content of the published article.

**Disclosure statement**

The authors declare that they have no known competing financial interests or personal relationships that could have appeared to influence the work reported in this paper.

**Data availability statement**

Raw point cloud data cannot be made available due to non-disclosure agreements with mine site. Other research data will be made available upon reasonable request.

**Funding**

No funding was received.